\pdfoutput=1

\documentclass[11pt,a4paper]{article}
\usepackage[hyperref]{emnlp2020}
\usepackage{times}
\usepackage{latexsym}

\usepackage{microtype}
\usepackage{todonotes}
\usepackage{tablefootnote}
\usepackage{appendix}
\usepackage[capitalize]{cleveref}

\aclfinalcopy 
\def\aclpaperid{2078} 

\newcommand{\roberta}{{RoBERTa}}
\newcommand{\wikipedia}{{Wikipedia}}

\title{Multi-Stage Pre-training for Low-Resource Domain Adaptation}

\author{Rong Zhang\thanks{\hspace{2mm}Both authors contributed equally.}, Revanth Gangi Reddy\footnotemark[1]\hspace{1.5mm}\thanks{\hspace{2mm}Work done during AI Residency at IBM Research.}, Md Arafat Sultan, Vittorio Castelli, Anthony Ferritto,\\ \textbf{Radu Florian, Efsun Sarioglu Kayi, Salim Roukos, Avirup Sil\thanks{\hspace{2mm}Corresponding author.}, Todd Ward} \\
  IBM Research AI \\ 
  \texttt{\{zhangr,avi,raduf,roukos,toddward,vittorio\}@us.ibm.com},\\
  \texttt{g.revanthreddy111@gmail.com}, \texttt{\{arafat.sultan,aferritto\}@ibm.com},\\
  \texttt{efsun@gwu.edu}}

\date{}

\begin{document}
\maketitle
\begin{abstract}

Transfer learning techniques are particularly useful in NLP tasks where a sizable amount of high-quality annotated data is difficult to obtain. Current approaches directly adapt a pre-trained language model (LM) on in-domain text before fine-tuning to downstream tasks. We show that extending the vocabulary of the LM with domain-specific terms leads to further gains. 
To a bigger effect, we utilize structure in the unlabeled data to create auxiliary synthetic tasks, which helps the LM transfer to downstream tasks.
We apply these approaches incrementally on a pre-trained Roberta-large LM and show considerable performance gain on three tasks in the IT domain: Extractive Reading Comprehension, Document Ranking and Duplicate Question Detection.

\end{abstract}

\section{Introduction}

Pre-trained language models 
\cite{radford2019language, devlin-etal-2019-bert, liu2019roberta} have pushed performance in many natural language processing
tasks to new heights. The process of model construction has effectively been reduced to extending the pre-trained LM architecture with simpler task-specific layers, while fine-tuning on labeled target data.
In cases where the target task has limited labeled data, prior work has also employed transfer learning by pre-training on a source dataset with abundant labeled data before fine-tuning on the target task dataset \cite{min-etal-2017-question, chung-etal-2018-supervised, wiese-etal-2017-neural-domain}.
However, directly fine-tuning to a task in a new domain may not be optimal when the domain is distant in content and terminology from the pre-training corpora.

To address this language mismatch problem, recent work \cite{alsentzer-etal-2019-publicly, Lee_2019, beltagy-etal-2019-scibert, gururangan2020dont} has adapted pre-trained LMs to specific domains by continuing to train the same LM on target domain text. 
Similar approaches are also used in multilingual adaptation, where the representations learned from multilingual pre-training are further optimized for a particular target language \cite{liu2020multilingual, bapna-firat-2019-simple}.
However, many specialized domains contain their own specific terms that are not part of the pre-trained LM vocabulary. Furthermore, in many such domains, large enough corpora may not be available to support LM training from scratch.
To resolve this out-of-vocabulary issue, in this work, we \emph{extend} the open-domain vocabulary with in-domain terms while adapting the LM, and show that it helps improve performance on downstream tasks.

While language modeling can help the model better encode the domain language, it might not be  sufficient to gain the domain knowledge necessary for the downstream task. We remark, however, that such unlabeled data in many domains can have implicit structure which can be taken advantage of.
 For example, in the IT domain, technical documents are often created using predefined templates, and support forums have data in the form of questions and accepted answers. In this work, we propose to make use of the structure in such unlabeled domain data to create synthetic data that can provide additional domain knowledge to the model. Augmenting training data with generated synthetic examples has been found to be effective in improving performance on low-resource tasks.
\citeauthor{golub-etal-2017-two}~\shortcite{golub-etal-2017-two}, \citeauthor{yang-etal-2017-semi}~\shortcite{yang-etal-2017-semi}, \citeauthor{lewis-etal-2019-unsupervised}~\shortcite{lewis-etal-2019-unsupervised} and \citeauthor{dhingra-etal-2018-simple}~\shortcite{dhingra-etal-2018-simple} develop approaches to generate natural questions that can aid downstream question answering tasks. However, when it is not possible to obtain synthetic data that exactly fits the target task description, we show that creating auxiliary tasks from such unlabeled data can be useful to the downstream task in a transfer learning setting. 
 

For preliminary experiments in this short paper, we select the IT domain, partly because of the impact such domain adaptation approaches can have in the technical support industry.
The main contributions of this paper are as follows: \textbf{(1)} We show that it is beneficial to extend the vocabulary of a pre-trained language model while adapting it to the target domain. 
\textbf{(2)} We propose to use the inherent structure in unlabeled data to formulate synthetic tasks that can transfer to downstream tasks in a low-resource setting.
\textbf{(3)} In our experiments, we show considerable improvements in performance over directly fine-tuning an underlying \roberta-large LM \cite{liu2019roberta} on multiple tasks in the IT domain: extractive reading comprehension (RC), document ranking (DR) and duplicate question detection (DQD).\footnote{Scripts are available  \href{https://github.com/IBM/techqa/tree/master/synthetic}{here}.}

\section{Datasets}
 We use two  publicly available IT domain datasets.  Table~\ref{tab:Datasets} shows their size statistics.
 \\
 
 \noindent\textbf{TechQA} \cite{castelli2019techqa} is an extractive reading comprehension \cite{rajpurkar2016squad} dataset developed from real user questions in the customer support domain. Each question is accompanied by 50 documents, at most one of which has the answer. A companion collection of 801K unlabeled Technotes is provided to support LM training. In addition to the primary reading comprehension task (TechQA-RC), we also evaluate on a new document ranking task (TechQA-DR). Given the question, the task is to find the document that contains the answer.\\

 \noindent\textbf{AskUbuntu}\footnote{\href{https://askubuntu.com}{\textit{askubuntu.com}}} 
 \cite{lei-etal-2016-semi} is a dataset containing user-marked pairs of similar questions from Stack Exchange\footnote{\href{https://stackexchange.com/}{\textit{stackexchange.com}}}, which was developed for a duplicate question detection task (AskUbuntu-DQD).
 A static offline dump of AskUbuntu, which is organized as a set of forum posts\footnote{\href{https://archive.org/download/stackexchange/askubuntu.com.7z}{\textit{archive.org/download/stackexchange/askubuntu.com.7z}}}, is also available  and can be used
 for LM training.


\begin{table}
\centering
\begin{tabular}{lcccc}
\hline \textbf{Dataset} & \textbf{Train} & \textbf{Dev} & \textbf{Test} & \textbf{Unlabeled}\\ \hline
TechQA & 600 & 310 & 490 & 306M \\
AskUbuntu & 12,724 & 200 & 200 & 126M \\
\hline
\end{tabular}
\caption{Size statistics for two IT domain datasets.
Train/Dev/Test: $\#$ examples, Unlabeled: $\#$ tokens.}
\label{tab:Datasets}
\end{table}

\section{Vocabulary Extension for LM Adaptation}
Texts in specialized fields including technical support in the IT domain may contain numerous technical terms which are not found in open domain corpora and are therefore not well captured by the vocabulary of out-of-the-box LMs. These terms are often over-segmented into small pieces (sub-word tokens) by the segmenter rules, which are learned from the statistics of open domain language.

As an example, the token out-of-vocabulary (OOV) rate of the standard \roberta\ vocabulary in the TechQA Technotes data is 19.8\% 
and the BPE/TOK ratio is 
1.32.
Contrast this with the analogous figures for 1M randomly selected \wikipedia\ sentences,
where the OOV rate is only 8.1\% and the BPE/TOK ratio is 1.12. While transformer-based pre-trained language models \cite{devlin-etal-2019-bert,liu2019roberta} yield better representations of previously unseen tokens than traditional $n$-gram models, over-segmentation can still cause 
degradation in downstream task performance.  

We address this challenge by augmenting the vocabulary of the pre-trained LM with 
frequent in-domain words.
Specifically, the most frequent OOV tokens after tokenization are recorded and used to  
bypass the BPE segmentation stage.  This prevents the segmenter from splitting these terms into smaller pieces. New entries in the LM vocabulary and corresponding word embeddings are created for these tokens. In our experiments, the 
number of such protected tokens is decided using an empirical criterion: 
we require that 95\% 
of the in-domain data be covered by the extended vocabulary.  We add 10k 
new items
to the vocabulary for the Technotes corpus and 5k for the AskUbuntu corpus. The variation in coverage due to different numbers of new vocabulary entries is shown in the appendix. The pre-trained LM is then adapted to the domain-specific corpus via masked LM (MLM) training. The embeddings of the new vocabulary are randomly initialized and then learned during the MLM training. The embeddings of existing vocabulary are also fine-tuned in this phase.

\section{Task-Specific Synthetic Pre-training}
\label{section:synt}

While in-domain LM pre-training reveals novel linguistic patterns in target domain text, in many domains including technical documents, structure present in unlabeled text can contain useful information closer to actual end tasks.
In this section, we propose to utilize such structure in unlabeled data to create auxiliary pre-training tasks and associated synthetic training data, which in turn can help target tasks via transfer learning.\newline 

\noindent \textbf{TechQA.}
The TechQA dataset release contains a companion Technotes collection with 801K human written documents with titled sections. 
We observe that certain sections in these documents (e.g., \textit{Abstract}, \textit{Error Description} and \textit{Question}) correspond to a problem description, while others (e.g., \textit{Cause} and \textit{Resolving the Problem}) describe the solution\footnote{Here is a sample technote: \href{https://leaderboard.techqa.us-east.containers.appdomain.cloud/examples/swg21318593.html}{Link}}.
We create an auxiliary reading comprehension (RC) task from these documents.
Specifically, if a document contains both problem and solution sections, a synthetic example is created where 
the problem description section is the query, the solution section is the target answer, and the entire document excluding the query section is the context.
Additionally, ten other documents are sampled from the Technotes corpus as negatives to simulate \textit{unanswerable} examples.
This auxiliary task trains an intermediate RC model which predicts the start and end positions of the solution section as the answer given the document and the problem description.
While our main goal here is to generate long-answer examples common in TechQA, the general idea of utilizing the document structure can be applicable in other scenarios including in scientific domains like 
Bio/Medical~\cite{tsatsaronis-2015,Lee_2019} where structured text is relatively common.\newline

\noindent \textbf{AskUbuntu.}
The AskUbuntu dataset contains a web dump of forum posts, each containing a question and multiple answers, with one answer possibly labeled by users as ``Accepted". Motivated by \cite{conf-ijcai-QiuH15, lei-etal-2016-semi, ruckle-etal-2019-neural}, we create an auxiliary answer selection task from this structure.
Each instance in the synthetic data for this task contains a question, its accepted answer as the positive class,
and an answer randomly sampled from other question posts as the negative class. 
An intermediate classification model is learned from these annotations, whose weights are used to initialize the target duplicate question detection (DQD) model.
Even though this auxiliary task adopts a different question-answer classification objective than the DQD task's objective of question-question classification, our experimental results show that the former still serves a good initialization for the latter.

\section{Experiments}

\begin{table*}[h]
\centering
\begin{tabular}{l|cc|cc}
\hline
 \textbf{Model}& \multicolumn{2}{c|}{\textbf{Dev}} & \multicolumn{2}{c}{\textbf{Test}} \\
 
   & \textbf{HA\_F1} & \textbf{F1} &  \textbf{HA\_F1} & \textbf{F1} \\
   \hline
   BERT  & 34.7 & 55.3 & 25.4 & 53.0 \\
\hline
  RoBERTa & 35.0 \small(1.6) & 58.2 \small(1.1) & 29.3  & 54.4  \\
\hline
  + Domain LM & 35.2 \small(1.5) & 58.3 \small(1.7) & - & -  \\
   + 10k Vocab Ext. & 36.9 \small(1.7) & 58.5 \small(0.7)& - & -  \\
    + RC Pre-training & 39.7 \small(1.2) & 59.0 \small(0.7)& - & -\\
     + Data Augmentation & \textbf{40.6} \small(1.4)& \textbf{59.9} \small(1.0) & \textbf{32.1}  & \textbf{56.7} \\
   \hline
\end{tabular}
\caption{Results on TechQA-RC task. Each row with a + adds a step to the previous row. HA\_F1 refers to F1 for answerable questions. Numbers in parentheses show standard deviation.}
  \label{tab:QA}
\end{table*}

\begin{table*}[h]
\centering
\begin{tabular}{l|cc|cc}
\hline
 \textbf{Model}& \multicolumn{2}{c|}{\textbf{Dev}} & \multicolumn{2}{c}{\textbf{Test}
 } \\

    & \textbf{M@1} & \textbf{M@5}& \textbf{M@1} & \textbf{M@5} \\
   \hline
  IR & 0.437 & 0.637 & - & - \\

  RoBERTa & 0.576 \small(0.017) & 0.770 \small(0.027) & 0.512  & 0.748 \\
\hline
  + Domain LM  & 0.593 \small(0.020) & 0.808 \small(0.021) & - & - \\

   + 10k Vocab Ext.  & 0.596 \small(0.013) & 0.790 \small(0.024) & - & - \\

    + RC Pre-training & 0.625 \small(0.014) & 0.826 \small(0.023) & - & -\\

    + Data Augmentation & \textbf{0.638} \small(0.029)  & \textbf{0.850} \small(0.012)& \textbf{0.536} & \textbf{0.808} \\
    \hline
\end{tabular}
\caption{Experimental results on TechQA-DR task. Each row with a + adds a step to the previous row. M@1 is short for Match@1 and M@5 for Match@5. Numbers in parentheses show standard deviation.}
  \label{tab:DR}
\end{table*}

\begin{table*}[t]
\centering
\begin{tabular}{l|cccc|cccc}
\hline
   \textbf{Model} & \multicolumn{4}{c|}{\textbf{Dev}} & \multicolumn{4}{c}{\textbf{Test}} \\
 
  &\textbf{MAP} & \textbf{MRR} & \textbf{P@1} &\textbf{P@5} & \textbf{MAP} & \textbf{MRR} & \textbf{P@1} & \textbf{P@5} \\
  \hline
  RoBERTa & 0.634  & 0.733  & 0.588  & 0.514  & 0.663 & 0.778  & 0.654  & 0.510 \\
  
  & \small(0.009) & \small(0.014) & \small(0.022) & \small(0.010) & \small(0.014) & \small(0.024) & \small(0.038) & \small(0.009) \\
   \hline
   
   + Domain LM & 0.647 & 0.753 & 0.622 & 0.523 & 0.677 & 0.799 & 0.676 & 0.515 \\
   
    & \small(0.007) & \small(0.021) & \small(0.029) & \small(0.009) & \small(0.012) & \small(0.019) & \small(0.028) & \small(0.008) \\
  \hline
 
   + 5k Vocab Ext. & 0.653 & 0.750 & 0.608 & 0.532 & 0.686 & 0.817 & 0.704 & 0.517 \\
   
      & \small(0.016) & \small(0.024) & \small(0.038) & \small(0.012) & \small(0.017) & \small(0.020) & \small(0.033) & \small(0.004) \\

  \hline

   + DQD Pre-training & \textbf{0.672} & \textbf{0.775} & \textbf{0.647} & \textbf{0.548} & \textbf{0.704} & \textbf{0.825} & \textbf{0.714} & \textbf{0.532} \\

     &  \small(0.008) & \small(0.012) & \small(0.023) & \small(0.007) & \small(0.012) & \small(0.015) & \small(0.028) & \small(0.008) \\

   \hline
\end{tabular}
\caption{Experimental results on AskUbuntu-DQD task. Each row with a + adds a step to the previous row. P@1 and P@5 refer to Precision@1 and Precision@5, respectively. Numbers in parentheses show standard deviation.}
 \label{tab:AskUbuntu}
\end{table*}

\subsection{Setup}


Our experiments build on top of the \roberta-large LM.
We adopt the standard methodology of using the pre-trained LM as the encoder and processing the contextualized representations it produces using task-specific layers. For the TechQA-RC task, we follow~\cite{devlin-etal-2019-bert} and predict the start and end position of the answer span with two separate classifiers, trained using cross entropy loss. For the TechQA-DR and AskUbuntu-DQD tasks, we follow~\cite{adhikari2019docbert} and classify the [CLS] token representation at the final layer with a binary classifier trained using the binary cross entropy loss; during inference, we rank the documents or questions according to their classification score. 
For all the tasks, during finetuning, we train the entire model end-to-end. We refer the reader to the appendix for details on hyperparameter values for all the experiments.

For the TechQA-RC task, we report both the main metric, F1, and the ancillary F1 for answerable questions, HA\_F1, to capture the effects of our approach both on the end-to-end pipeline (F1) and on the answer extraction component (HA\_F1). For TechQA-DR, models are evaluated by Match@1 and Match@5. For AskUbuntu-DQD, we report MAP, MRR, Precision@1 and Precision@5 following~\cite{lei-etal-2016-semi}.



\subsection{Synthetic Pre-training Corpus and Labeled Data Augmentation}

Using the method described in section \ref{section:synt}, we use the 801K Technotes to construct a synthetic corpus for the TechQA tasks. The synthetic data contains 115K positive examples, each of which has 10 randomly selected documents as negatives. For the AskUbuntu-DQD, a 210K-example synthetic corpus is constructed from the web dump data, with a positive:negative example ratio of 1:1.

Since TechQA is a very-low resource dataset with only 600 training examples, we additionally apply data augmentation techniques to increase the size of the training set. We use simple data perturbation strategies, such as adding examples with only parts of the original query, randomly dropping words in query and passage, duplicating positive examples, removing stop words, dropping document title in the input sequence etc., to increase the size of the training set by 10 times. 
This augmented training set is only used under the
data augmentation setting while fine-tuning on the TechQA tasks.


\subsection{Results and Analysis}

For each of our approaches, we show performance of the model when fine-tuned on the downstream tasks in TechQA and AskUbuntu datasets. All the numbers reported are averages 
over 5 seeds, 
unless otherwise stated.
Standard deviation numbers are shown in parentheses.
\newline

\noindent \textbf{TechQA-RC} 
Table~\ref{tab:QA} describes the performance on the RC task in the TechQA dataset. 
The BERT baseline numbers are from~\cite{castelli2019techqa}.
Here, model performance is compared on the dev set and we report the blind test set numbers\footnote{Obtained by submitting to the \href{https://leaderboard.techqa.us-east.containers.appdomain.cloud}{\textit{TechQA leaderboard}}.} for our single-best baseline and final models.


Adapting the LM without extending the vocabulary yields just 0.2 points over the \roberta-large baseline. Augmenting the vocabulary by 10k word pieces improves the HA\_F1 score by 1.7 points. Furthermore, our RC-style synthetic pre-training yields a considerable improvement of 2.8 points on HA\_F1 and 0.5 points on F1. 
Finally,  data augmentation further boosts performance by about a point on both HA\_F1 and F1, suggesting that data augmentation via simple perturbations \ can be effective in a very-low resource setting. \newline

\noindent \textbf{TechQA-DR} 
Table \ref{tab:DR} shows results from our experiments on the auxillary document ranking task over the TechQA dataset \footnote{Since this is not the official task in the TechQA dataset, numbers on the test set were obtained by the TechQA leaderboard manager who agreed to run our scoring script on an output file produced by our submission}. 
We use BM25 \cite{articleBM25} as our IR baseline.
We see that the 
RoBERTa
models substantially outperform the IR system. Although vocabulary expansion only helps by 0.3 points in  Match@1, 
we see considerable improvements in performance from our other approaches.
The ``RC Pre-training" entry shows a Match@1 improvement of 2.9 points over the language modelling. This demonstrates the effectiveness of pre-training on an ancillary task in a transfer-learning setting for the document ranking task.
We further see an improvement of 1.3 points from data augmentation.

~\newline
\noindent \textbf{AskUbuntu-DQD} 
Table \ref{tab:AskUbuntu} shows results for the 
DQD
task on the AskUbuntu dataset. We see that our methods give incremental improvements in performance. Our final model is considerably better than the \roberta-large baseline on all four metrics. We see the biggest gain in performance from the synthetic pre-training task demonstrating its relevance to the DQD task.  For this dataset, we didn't explore data augmentation strategies because  it had a considerable number of training instances (see Table~\ref{tab:Datasets}) compared to the TechQA dataset. 


\section{Conclusion}
In this work, 
we show that it is beneficial to extend the vocabulary of the LM while 
fine-tuning it on the target domain language. We show that 
extending the pre-training
with task-specific synthetic 
data is an effective domain adaptation strategy. We empirically demonstrate that structure in the unsupervised domain data can be used to formulate auxillary pre-training tasks that can help downstream low-resource tasks like question answering and document ranking.
In our preliminary experiments, we empirically show considerable improvements in performance
over a standard RoBERTa-large LM on multiple tasks.
In future work, we aim to extend our approach to more domains and explore more generalizable approaches for unsupervised domain adaptation.



\section*{Acknowledgments}

We thank Cezar Pendus for his help with submitting to the TechQA leaderbaord. We would also like to thank the multilingual NLP team at IBM Research AI and the anonymous reviewers for their helpful suggestions and feedback.

\bibliographystyle{acl_natbib}
\bibliography{emnlp2020}

\clearpage
\newpage
\appendix 

\section{Appendix}

\subsection{Implementation Details}
In our experiments, we used the Fairseq toolkit~\cite{ott2019fairseq} for language modelling and the  Transformers library~\cite{Wolf2019HuggingFacesTS} for downstream tasks. 
For all of our target models, when fine-tuning on the downstream task, we choose the hyperparameters by grid search and pick the best models on the dev set according to the evaluation metrics for the corresponding task. For TechQA-RC task, we pick the best model according to (HA\_F1 + F1) and for TechQA-DR, we choose based on Match@1. For the AskUbuntu-DQD, we pick the best model based on MAP. The best hyperparamters for each of the tasks are shown in the \crefrange{tab:A_LM}{tab:A_DQD} below:

\begin{table}[h]
\small
\centering
\begin{tabular}{l|c}
\textbf{Hyperparameter} & \textbf{Setting} \\ 
\hline
WARMUP UPDATES & 10000 \\
PEAK LR & 0.00015 \\
TOKENS PER SAMPLE & 512 \\
MAX POSITIONS & 512 \\
MAX SENTENCES & 8 \\
UPDATE FREQ & 64  \\
OPTIMIZER & adam \\
DROPOUT &  0.1 \\
ATTENTION DROPOUT & 0.1  \\
WEIGHT DECAY & 0.01 \\
MAX Epochs & 5 \\
CRITERION & mask-whole-words \\
\hline
\end{tabular}
\caption{Hyperparameters for the LM training.}
\label{tab:A_LM}
\end{table}

\begin{table}[h]
\small
\centering
\begin{tabular}{l|c}
\textbf{Hyperparameter} & \textbf{Setting} \\ 
\hline
Learning Rate & 5.5e-6 \\
Max Epochs & 15\\
Batch Size & 32\\
Max Sequence Length & 512\\
Document Stride & 192\\
Sampling Rate for Unanswerable Spans & 0.15\\
Maximum Query Length & 110\\
Maximun Answer Length & 200\\
\hline
\end{tabular}
\caption{Hyperparameters for the TechQA-RC task.}
\label{tab:A_RC}
\end{table}

\begin{table}[h]
\small
\centering
\begin{tabular}{l|c}
\textbf{Hyperparameter} & \textbf{Setting} \\ 
\hline
Learning Rate & 2.5e-6 \\
Max Epochs & 20\\
Batch Size & 32\\
Max Sequence Length & 512\\
Document Stride & 192 \\
Sampling Rate for Negative Documents & 0.1\\
Maximum Query Length & 110\\
\hline
\end{tabular}
\caption{Hyperparameters for the TechQA-DR task.}
\label{tab:A_DR}
\end{table}

\begin{table}[h]
\small
\centering
\begin{tabular}{l|c}
\textbf{Hyperparameter} & \textbf{Setting} \\ 
\hline
Learning Rate & 5.5e-6 \\
Max Epochs & 5\\
Batch Size & 32\\
Max Sequence Length & 512\\
Maximum Question Length & 256\\
Maximun Answer Length & 256\\
\hline
\end{tabular}
\caption{Hyperparameters for the AskUbuntu-DQD task.}
\label{tab:A_DQD}
\end{table}

\newpage
\subsection{Extension of Vocabulary}

The Table \ref{tab:A_EXT} below shows the variation of coverage and BPE/TOK ratio with the number of word pieces added to the vocabulary for the Technotes Collection.

\begin{table}[h]
\centering
\small
\begin{tabular}{ccc}
\hline \textbf{\# of Added Word Pieces} & \textbf{Coverage} & \textbf{BPE/TOK}\\ \hline
+0k & 80.2\% & 1.32 \\
+5k & 94.4\% & 1.13 \\
+10k & 95.4\% & 1.11 \\
+15k & 95.8\% & 1.10 \\
\hline
\end{tabular}
\caption{Coverage and BPE/TOK ratio vs the number of word pieces added to the vocabulary for the Technotes collection.}
\label{tab:A_EXT}
\end{table}


   




\newpage


\end{document}